\documentclass{article}
\pdfoutput=1
\usepackage{microtype}
\usepackage{graphicx}
\usepackage{subfigure}
\usepackage{booktabs} 

\usepackage{hyperref}

\usepackage{multirow}
\usepackage{booktabs}       
\usepackage{amsmath}
\usepackage{amsfonts}
\usepackage{xcolor}

\newcommand{\BF}[1]{\textbf{#1}}
\newcommand{\IT}[1]{\textit{#1}}

\newcommand{\etal}{\textit{et al}.}

\usepackage[accepted]{icml2021}

\icmltitlerunning{Overwrite Quantization}

\begin{document}

\twocolumn[
\icmltitle{OverQ: Opportunistic Outlier Quantization \\ for Neural Network Accelerators
}



\icmlsetsymbol{equal}{*}

\begin{icmlauthorlist}
\icmlauthor{Ritchie Zhao*}{c,m}
\icmlauthor{Jordan Dotzel*}{c}
\icmlauthor{Zhanqiu Hu}{c}
\icmlauthor{Preslav Ivanov}{c}
\icmlauthor{Christopher De Sa}{c}
\icmlauthor{Zhiru Zhang}{c}
\end{icmlauthorlist}

\icmlaffiliation{c}{Cornell University}
\icmlaffiliation{m}{Microsoft}

\icmlcorrespondingauthor{Jordan Dotzel}{dotzel@cornell.edu}

\icmlkeywords{Machine Learning, Quantization, Hardware}

\vskip 0.3in
]



\printAffiliationsAndNotice{\icmlEqualContribution} 
\begin{abstract}

Outliers in weights and activations pose a key challenge for fixed-point quantization of neural networks. 
While they can be addressed by fine-tuning, this is not practical for ML service providers (e.g., Google or Microsoft) who often receive customer models without training data. 
Specialized hardware for handling activation outliers can enable low-precision neural networks, but at the cost of nontrivial area overhead. 
We instead propose overwrite quantization (OverQ), a lightweight hardware technique that opportunistically increases bitwidth for activation outliers by overwriting nearby zeros. It has two major modes of operation: range overwrite and precision overwrite. Range overwrite reallocates bits to increase the range of outliers, while precision overwrite reuses zeros to increase the precision of non-outlier values.
Combining range overwrite with a simple cascading logic, we handle the vast majority of outliers to significantly improve model accuracy at low bitwidth. Our experiments show that with modest cascading, we can consistently handle over 90\% of outliers and achieve +5\% ImageNet Top-1 accuracy on a quantized ResNet-50 at 4 bits.
Our ASIC prototype shows OverQ can be implemented efficiently on top of existing weight-stationary systolic arrays with small area increases per processing element. We imagine this technique can complement modern DNN accelerator designs to provide small increases in accuracy with insignificant area overhead.

\end{abstract}
\begin{figure}[htb]
  \begin{center}
    \hfill
    \begin{minipage}{0.5\textwidth}
      \centering
      \includegraphics[width=\columnwidth]{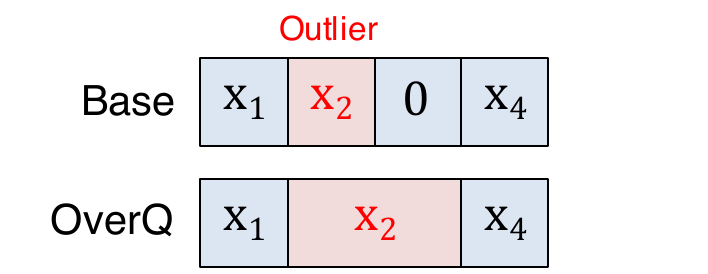}
    \end{minipage}
    \hfill
    \vspace{-0.05in}
    \caption[Basic idea of OverQ)]{\BF{Overwrite Quantization (OverQ) --- }
    Outlier $x_{i}$ can overwrite the adjacent value $x_{i+1}$ when $x_{i+1}$ is zero.
    If so, $x_{i}$ uses twice the bitwidth to extend either its range or precision.
    }
    \label{fig:basic}
  \end{center}
\vspace{-0.1in}
\end{figure}
\begin{figure*}[ht]
  \begin{center}
    \hfill
    \begin{minipage}{0.5\textwidth}
      \centering
      \includegraphics[width=\columnwidth]{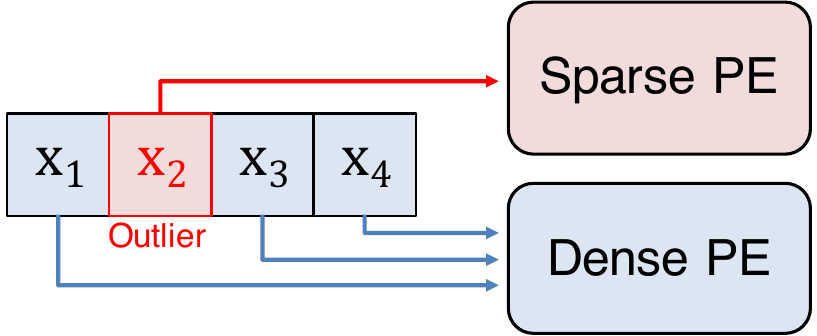}
    \end{minipage}
    \hfill
    \vspace{-0.05in}
    \caption[OLAccel]{\BF{OLAccel~\cite{park2018outlier} --- }
    OLAccel: Uses twice the bitwidth to represent the outliers with a separate hardware accelerator. The remaining elements are processed by a dense PE.
    }
    \label{fig:olaccel}
  \end{center}
\vspace{-0.1in}
\end{figure*}

\section{Introduction}
\label{sec-overq-intro}

Deep neural networks (DNNs) have achieved state-of-the-art results in many machine learning domains including computer vision, natural language processing, and robotic control. 
However, DNNs are limited by their high computational and storage requirements. For example, the cutting-edge GPT-3 family of models~\cite{brown2020gpt3} contains up to 175 billion parameters and requires hundreds of GiB for storage.
These high execution costs impede DNN deployment on edge devices~\cite{xu2018scaling} and have measurable impact on the the carbon footprint of datacenters~\cite{strubell2019energy}.

Quantizing neural network weights and activations from floating-point to low-precision fixed-point is a promising approach for reducing DNN size and complexity.
DNN quantization is a highly active research topic, with many works proposing to fine-tune the models during training to make them quantization-friendly~\cite{jacob2018quantization,yang2018synetgy,choi2018pact,dong2020hawqv2,banner2018scalable}.
Recently, however, a number of works have argued for the importance of post-training quantization techniques~\cite{tensorrt2017slides,zhao2019ocs,nagel2019equalization,meller2019same} and data-free quantization \cite{cai2020zeroq, li2020dfqf}. 
In these methods, the models are fully-trained and the quantization parameters are tuned either with no data or a small profiling dataset. 
These techniques are important because ML service providers often must optimize their customers' neural networks without having access to the training data.

Among these methods, uniform quantization is the most popular since it simplifies the hardware and requires only efficient scalar multiplication and addition.
Yet, uniform quantization is highly sensitive to the existence of outliers, which directly affect the accuracy of the quantized model. 
Within most networks, values are distributed in a bell-shaped distribution, with the majority of values concentrated near zero and a long tail of rare outliers.
These outliers are often the most significant values, yet representing them accurately forces high quantization error on the small values.
A number of techniques have been proposed to deal with them including clipping~\cite{sung2015resiliency,shin2016fixed,mckinstry2018clip,tensorrt2017slides} and channel splitting~\cite{zhao2019ocs,park2019celldiv}.
These techniques have made post-training quantization of many popular DNNs possible to around 8 bits without accuracy degradation.

For even lower precision, specialized hardware can be used.
Park \etal~\cite{park2018value,park2018outlier} proposed an outlier-aware DNN accelerator (OLAccel) that uses a conventional processing engine (PE) for central activations and a second sparse PE for outliers (shown in Figure~\ref{fig:olaccel}).
OLAccel achieves less than $1\%$ accuracy loss on AlexNet using 4 bits for most values and 16 bits for a small percentage of outliers.
Though effective, OLAccel's major drawback is using an entirely separate outlier engine, which requires additional multiply-accumulate (MAC) units and incurs hardware overhead due to sparsity.

In this paper, we present overwrite quantization (OverQ), a post-training quantization technique that uses \IT{lightweight} architectural extensions to address outliers (shown in Figure~\ref{fig:basic}).
To achieve this, OverQ \IT{opportunistically} increases the range of outliers by allowing them to overwrite adjacent zeros. Additionally, non-outliers can use these adjacent zero values for increased precision. We further increase the number of outliers handled by introducing cascading, which allows values to sequentially overwrite one another. We demonstrate the value of this technique by mapping it to a weight-stationary systolic array and showing it significantly improves the accuracy on the majority of models, up to an increase of 5\% at low activation bitwidths.
An ASIC prototype built using high-level synthesis shows that OverQ has a light hardware footprint, making it suitable for being included on other DNN accelerator designs.

\begin{figure*}[th]
  \begin{center}
    \hfill
    \begin{minipage}{0.65\textwidth}
      \centering
      \includegraphics[width=\columnwidth]{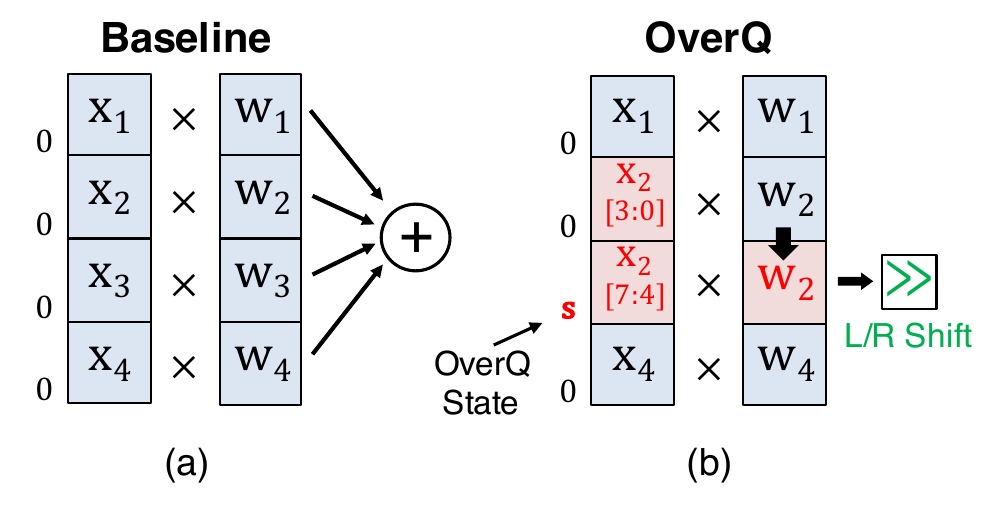}
    \end{minipage}
    \hfill
    \vspace{-0.05in}
    \caption[OverQ Dot Product]{\BF{OverQ Dot Product --- }
    (a) Standard multiply-and-accumulate (MAC) operation that makes up the majority of computation in DNNs
    (b) OverQ MAC: an outlier $x_{i}$ can overwrite the adjacent value $x_{i+1}$ when $x_{i+1}$ is zero.
    If so, twice the bitwidth is used to represent $x_{i}$. The weight $w_{i}$ is copied to the adjacent cell. The OverQ state determines the weight overwrite and shift-direction, if relevant, varying from one or two bits depending on which OverQ features are supported.
    }
    \label{fig:overq-dot-product}
  \end{center}
\vspace{-0.1in}
\end{figure*}

\section{Related Work}

We limit the discussion of related work to existing post-training quantization techniques in software and hardware and methods for dynamic precision. A few of these were the motivation for some of the core ideas laid our here. OLAccel~\cite{park2018outlier} motivated us to explore the separate treatment of outliers, and Precision Gating~\cite{zhang2020pg} motivated the usage of using a dual-precision inference. We expand on these techniques and distinguish our work in the following paragraphs.

\subsection{Software Techniques for Outliers}
Clipping weights and activations to pre-determined thresholds is the simplest technique for controlling the effects of outliers. During clipping, all values outside of clipping thresholds are rounded to those values. For weights, the values are immediately accessible while for activations, the values must be profiled using small datasets or artificial distributions.
These clipping methods include minimal mean-squared (MMSE) error~\cite{sung2015resiliency,shin2016fixed}, percentile of values~\cite{mckinstry2018clip}, and KL divergence~\cite{tensorrt2017slides}. Weight splitting takes nodes or channels containing outlier weights and duplicates them while dividing the weight in half~\cite{zhao2019ocs,park2019celldiv}. However, since splitting requires static information on the locations of outliers, it cannot be directly applied to activations~\cite{zhao2019ocs}, whose outliers are input-dependent. 

\subsection{Specialized Hardware for Outliers}
Software techniques have difficulties reaching simultaneous weight and activation quantization below 8-bit without significant accuracy degradation.
Specialized hardware can help overcome this barrier.
Park \etal~\cite{park2018value,park2018outlier} proposed OLAccel, an outlier-aware DNN accelerator, which uses a dense PE for most activations and a sparse PE for outliers (Figure~\ref{fig:olaccel}).
Increasing the bitwidth for outliers affords greater dynamic range yet incurs non-trivial hardware overhead. 
Although Park \etal ~did not specify the exact area of the outlier PE, we can observe that: (1) the outlier PE requires additional MAC units since it operates at a different bitwidth; (2) the sparse representation incurs storage overhead.
For each outlier, 32 additional bits are used for indices.
In contrast, OverQ seeks to handle outliers in a higher precision but in a more area-efficient manner.

\subsection{Dynamic Precision}
Some works use dynamic precision during inference to allocate extra bitwidth to important areas of the input. 
For example, Precision Gating~\cite{zhang2020pg} evaluates most features in low-precision but then uses a trained mask to update a portion of the features in higher precision. SeerNet~\cite{cao2019seernet} likewise proposes a method for calculating a mask using a quantized model, then making full-precision sparse updates to the result. These dual-precision methods differ from ours in part because we need no training phase, we select the precision at the activation level as opposed to channel-level, and we have smaller hardware overhead.  

\begin{figure*}[tbhp]
  \begin{center}
    \hfill
    \begin{minipage}{0.95\textwidth}
      \centering
      \includegraphics[width=\columnwidth]{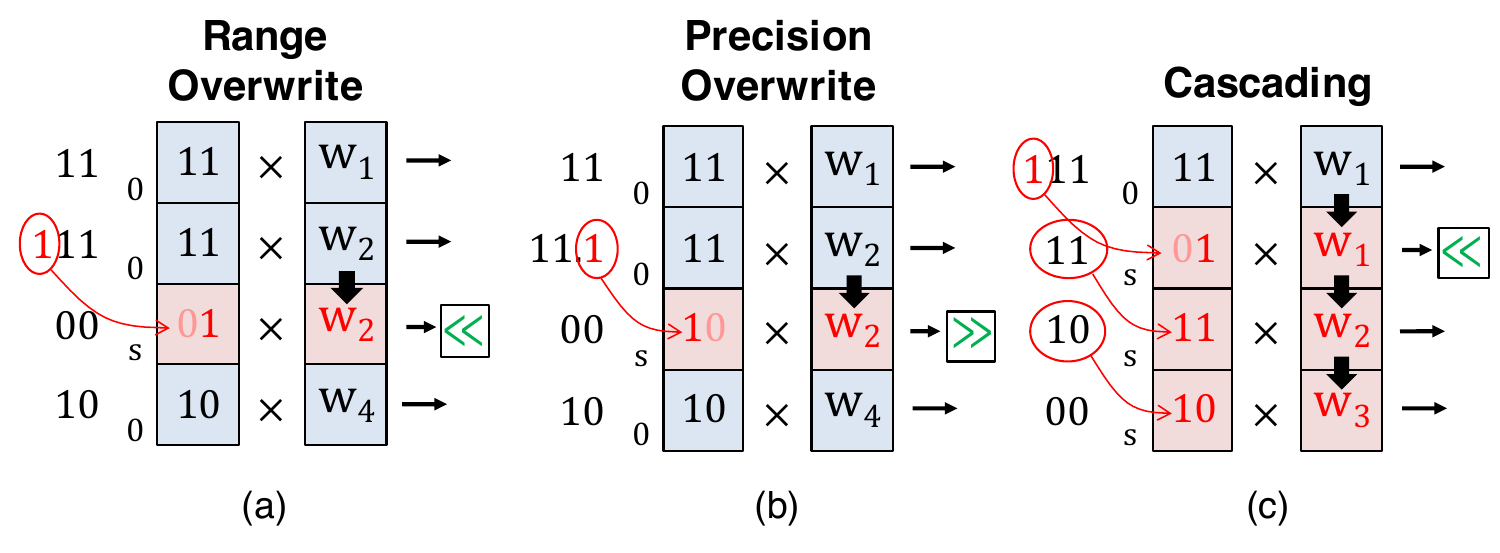}
    \end{minipage}
    \hfill
    \caption[OverQ Methods]{\BF{OverQ Methods
    --- }
    (a) If $x_{i}$ is an outlier, the adjacent slot stores out-of-range MSB bits.
    (b) If $x_{i}$ is not an outlier but $x_{i+1}$ is a zero, the adjacent slot can be ``reused'' to store out-of-range LSB bits of $x_{i}$.
    (c) If $x_{i}$ is an outlier and there is a zero within the cascade factor, $c$, then $x_{i}$ cascades the overwrites up to $x_{i+c}$. All intermediate values will be shifted. All methods require some shifting, except for the purely cascaded values, e.g. $w_2$ and $w_3$.
    }
    \label{fig:modes}
  \end{center}
\end{figure*}

\section{Overwrite Quantization}
OverQ dynamically extends the bitwidth for important values by overwriting nearby values.
OverQ exploits two properties of DNN activations: (1) outliers are rare but contribute disproportionately to a layer's output; (2) activation distributions contain many ReLU-induced zeros.
The first property suggests that overwrite occurs infrequently, limiting logic overhead.
The second property means that an outlier lies besides zeros with significant probability, although this depends heavily on the layer activation distribution and input. 

OverQ has two dominant modes, range overwrite (RO) and precision overwrite (PR). 
For range overwrite, outliers are defined as values that would be clipped using uniform quantization. Range overwrite uses adjacent values to extend the range for outliers. 
An outlier $x_{i}$ can \IT{overwrite} its adjacent value $x_{i+1}$ when $x_{i+1}$ is zero.
After the overwrite, $x_{i}$ is then represented with twice the normal range. Similarly, for precision overwrite, values can overwrite nearby zeros to increase their precision.

OverQ is applied along a single dimension of the activation tensor (width, height, or channels for convolutional networks).
Our experiments show that OverQ along the channels is more effective than along spatial dimensions, since adjacent channels exhibit less correlation than adjacent spatial locations in CNNs. 
This is a natural effect of the continuously varying spatial dimensions of model inputs. 
Correlation is an issue since there are no overwrite opportunities when zeros are adjacent to zeros, or outliers to outliers. 
For all our experiments, we apply OverQ along the input channel dimension.

\subsection{Computing with OverQ}
OverQ can be implemented as a lightweight extension on existing accelerator designs, and we now focus on the foundational unit for most linear operations in these accelerators. The dot product, shown in
Figure~\ref{fig:overq-dot-product}(a), between activations $x_{i}$ and weights $w_{i}$ is the basis of convolutional and fully-connected layers in DNNs.
Figure~\ref{fig:overq-dot-product}(b) shows the OverQ variant of the dot product.
If an overwrite is detected, OverQ routes the out-of-range bits of the value $x_{i}$ to overwrite $x_{i+1}$. These could be either additional more-significant bits (MSBs) for range overwrite or less-significant bits (LSBs) for precision overwrite. The slot for $x_{i+1}$ reads the OverQ state and copies $w_{i}$ to the position of $w_{i+1}$; the two products at indices $i$ and $i+1$ thus sum to $x_{i} \times w_{i}$ when appropriately shifted. Range overwrite requires a left shift to return the MSBs to their original positions, and precision overwrite requires a right shift.
Copying the weight is simple in hardware as long as we spatially unroll the vector such that the compute units for $x_{i}$ and $x_{i+1}$ are physically close.

The OverQ state encodes the overwrite status and shift direction. 
It needs a single-bit if only range overwrite is supported, which indicates whether the value is being overwritten. 
For the combination of range and precision overwrite, the state value requires two bits to distinguish among these states: range overwrite, precision overwrite, no OverQ.
At the circuit level, OverQ requires extra registers, interconnects, and logic, but since the area is dominated by multipliers, overall these increases are insignificant. This state variable requires a simple algorithm to be calculated. It can be integrated with existing logic in most accelerators to limit its overhead. Figure~\ref{fig:modes} gives numerical examples for both range and precision overwrite.

\begin{figure*}[thbp]
  \begin{center}
    \hfill
    \begin{minipage}{.9\textwidth}
      \centering
      \includegraphics[width=\columnwidth]{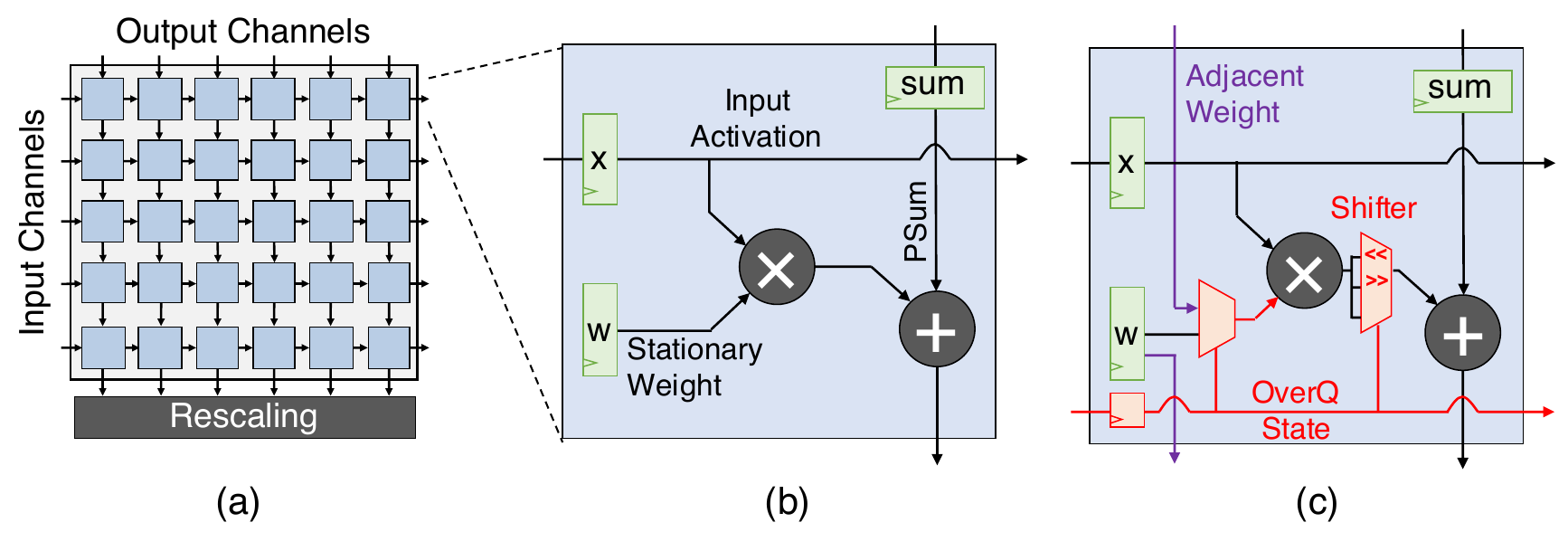}
    \end{minipage}
    \hfill
    \vspace{-0.05in}
    \caption[OverQ Hardware Architecture]{\BF{OverQ Hardware Architecture --- }
    OverQ can be supported with lightweight changes in a weight-stationary 2D spatial array, a common DNN accelerator design.
    (a) Baseline systolic array
    (b) Baseline PE showing weight, activation, and partial sum (PSum) along with MAC unit;
    (c) OverQ PE requires a state, muxes, and shifters.}
    \label{fig:overq-arch}
  \end{center}
\vspace{-0.1in}
\end{figure*}

\subsection{Cascading}

OverQ operates opportunistically by reallocating bitwidth from some values to others during inference. The opportunistic nature is important since it limits the amount of logic necessary to compute the OverQ state, and with the large size of activations in standard model layers, its results are still fairly predictable.
As shown in Figure~\ref{fig:modes}, this requires the overwriting values to be adjacent to zeros. 
Focusing on range overwrite, we quantify its effectiveness by \textit{outlier coverage}, which we define as the percentage of outliers handled by OverQ. We define an outlier as any value that the quantizer clips due to the restricted bitwidth.

There are many techniques that increase the outlier coverage.
For example, we can statically profile the activation distribution beforehand, note the channels with the most and least outliers, and re-index the channels before inference so that the channels with most outliers are next to those with most zeros.
This can increase the outlier coverage slightly on average; however, this requires a profiling dataset and ignores the input-dependent nature of the outliers.
Instead, we introduce a more powerful technique called cascading, which shifts a sequence of weights to allow overwriting near but non-adjacent values. 
This process is illustrated in Figure~\ref{fig:modes}(c). We control the extent of cascading through the \textit{cascade factor}, which determines the maximum length of the cascade. A cascade factor of 1, for example, would be the trivial case of looking at just the adjacent value, i.e. no cascading.
As with non-cascaded range overwrite, the outlier overwrites into the most adjacent cell, although with cascading this value now shifts into its adjacent cell, which shifts into its adjacent cell, and so forth.
This creates a cascade of overwrites using the existing weight sharing extensions of non-cascaded OverQ. 

Cascading significantly increases the outlier coverage by operating on a per-input basis. 
However, it requires extra logic to compute the appropriate states. 
Specifically, for $n$ outliers and a cascade factor of $c$, the simplest algorithm operates at $O(nc)$ since for of $n$ outliers it must look ahead $c$ values for zeros. We describe later in Section~\ref{sec:overq-arch} how we perform the OverQ state calculations in a realistic system, and we now demonstrate that $c$ is very small for standard models. This reduces the computation to roughly $O(n)$ in practice.

Cascading only needs to extend to the nearest zero, where all values in between will be overwritten and shifted. Therefore, an appropriate reference point is the expected distance from a given value to the nearest zero. Given the probability that a value is zero and assuming independent channel computations, $p_0$, the probability that a zero is within $c$ slots is:
\begin{equation}
\label{eq:coverage}
    P_k = 1 - (1 - p_0)^c
\end{equation}
This probability has a decaying exponential factor that implies higher cascade factors quickly become less important in increasing outlier coverage. Assuming models with batch norm and ReLU activations, this leads to half the activations being zero and approximately $p_0 = .5$. Table~\ref{tab:overq-coverage} shows the theoretical and empirical outlier coverage for several different cascade factors, $c$. These layers are taken from ResNet-50 and are quantized to 4 bits. As shown in the table, the number of zeros in a layer affects the outlier coverage, since more zeros on average would be adjacent to outliers. The outlier coverage has diminishing gains at higher cascade factors, as expected, although the total outlier coverage is higher than our simple model suggests. This comes from the assumption that adjacent values are statistically independent, where in reality these are coupled by sharing the same inputs.
\begin{table}[tbhp]
  \centering
  \small
  \caption[Cascading Outlier Coverage]{\BF{Cascading Outlier Coverage} -- the outlier coverage across cascade factors on arbitrary layers in ResNet-50. A cascade factor of 1 implies no cascading. `Theory' column is given by Equation \eqref{eq:coverage}.}
  \vspace{0.1in}
    \begin{tabular}{c|c|c|c|c}
    \toprule
    \BF{Cascade} & \BF{Theory} & \BF{Layer1} & \BF{Layer2} & \BF{Layer3}\\
    \BF{Factor}  &             &             &             &            \\
    \toprule
    1 & 50.0 & 64.1 & 56.9 & 47.8 \\ 		
    2 & 75.0 & 87.7 & 88.2 & 68.5 \\ 
    3 & 87.5 & 94.6 & 96.2 & 72.5 \\ 		
    4 & 93.8 & 96.1 & 98.4 & 79.7 \\ 	
    5 & 96.7 & 97.0 & 98.9 & 85.0 \\ 
    6 & 98.4 & 97.2 & 99.6 & 89.1 \\
    \bottomrule
    Zero Perc. & 50.0 & 51.1 & 69.1 & 30.3 \\
    \bottomrule
    \end{tabular}%
  \label{tab:overq-coverage}%
\end{table}%

\begin{figure*}[tbhp]
  \begin{center}
    \hfill
    \begin{minipage}{\textwidth}
      \centering
      \includegraphics[width=.9\columnwidth]{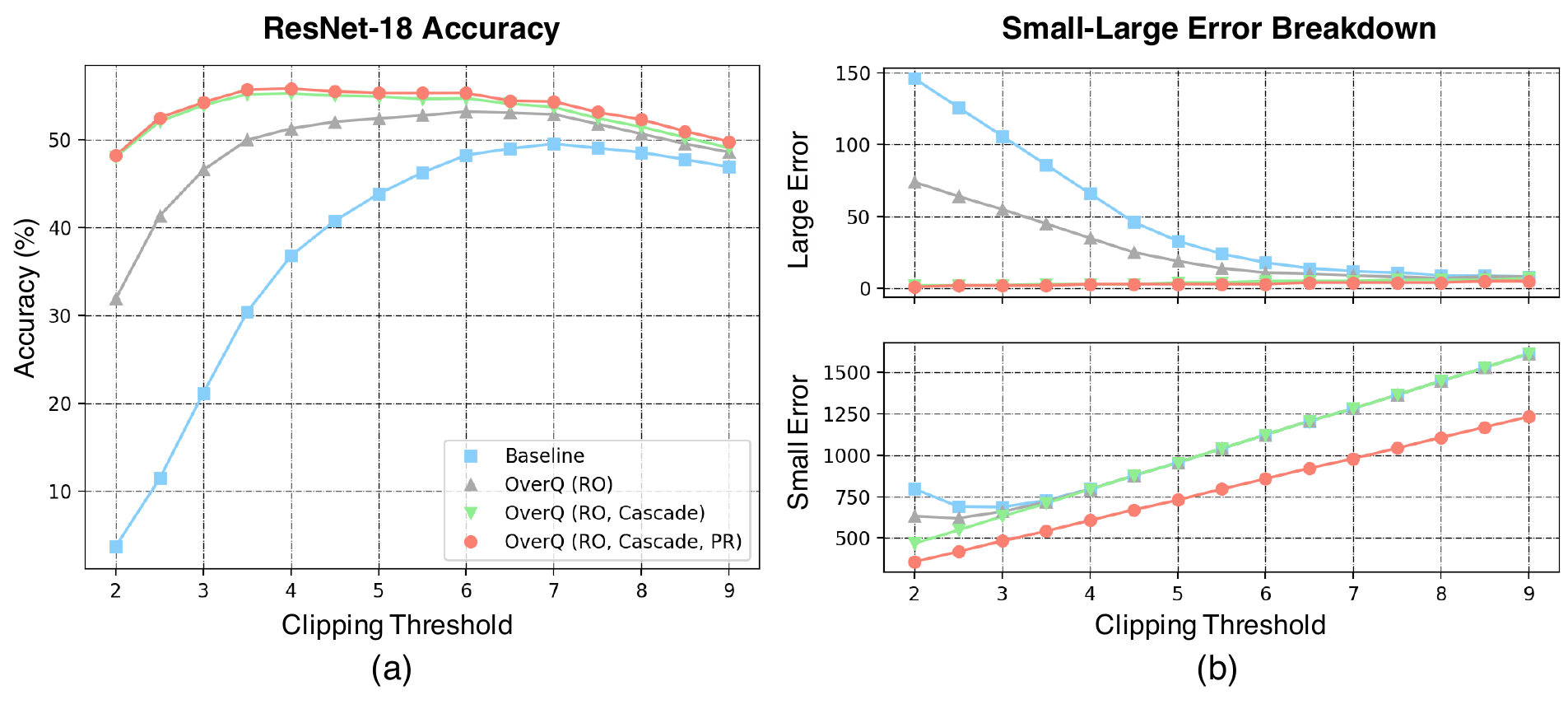}
    \end{minipage}
    \hfill
    \vspace{-0.05in}
    \caption[OverQ Analysis on ResNet-18]{\BF{OverQ Analysis on ResNet-18 --- }
    (a) Plot shows baseline quantization, range overwrite (RO), cascading, and the full OverQ method with precision overwrite (PR).
    Clip threshold is expressed as a multiple of the per-layer standard deviation. The model was quantized to 4-bit weights and activations.
    (b) Plots show the breakdown in quantization error on an arbitrary layer of ResNet-18 between small and large values. The split between small and large is 4 in this figure. Error is measured as the sum of all the absolute errors between the original and quantized distributions. 
    }
    \label{fig:overq-std}
  \end{center}
\vspace{-0.1in}
\end{figure*}

\section{Mapping OverQ to DNN Accelerators}
\label{sec:overq-arch}

From the outset, OverQ was designed for efficient hardware implementation in realistic DNN accelerators.
In this section, we show how OverQ can be added using only lightweight architectural changes to a \IT{weight-stationary} spatial array --- a common template for DNN accelerators.
Here, weight-stationary (WS) refers to a type of DNN dataflow~\cite{chen2017eyeriss} in which weights are held in processing elements (PEs) while inputs and partial sums move through the accelerator.
A recent study on dataflow choice in DNN hardware literature~\cite{yang2018dataflow} showed that weight-stationary dataflow was the most popular.
Experiments in the study also demonstrated that was the most hardware-efficient, albeit by only a small margin.

Weight-stationary spatial architectures, shown in Figure~\ref{fig:overq-arch}(a), are well-suited for OverQ.
The array spatially unrolls the input channels, which are mapped to rows along the vertical axis. 
Therefore, adjacent channels are mapped to physically adjacent PEs, which facilitates weight sharing during processing.
In more detail, Figure~\ref{fig:overq-arch}(a) shows how the input activations move from left to right while the output partial sums move from top to bottom. This systolic array contains many PEs, shown in Figure~\ref{fig:overq-arch}(b), where each contains a single multiplier and an adder. These architectures can map many types of MAC operations, including the 1x1 convolution that we prototype in hardware.

OverQ builds on top of the standard PE, as shown in Figure~\ref{fig:overq-arch}(c) by adding muxes, shifters, and registers.
The registers are necessary to propagate the OverQ state along with the activations during computation. The shifters must decide between range overwrite, precision overwrite, as well as being inactive for cascading. The muxes are responsible for the correct decoding of the OverQ state necessary to select to the proper datapath.

The OverQ state must be computed for the activations before they enter the systolic array.
This process must be done in the accumulation and rescaling unit, since only here the outputs are temporarily in higher precision.
This unit is modified to determine where OverQ should operate and which mode it should select.
Despite this logic, the dominant resource overhead will be from the modifications to each PE since they scale at a faster rate than the rescaling unit. The rescaling unit scales with the width of the array since each column has its own accumulator and scale factors. The PEs, on the other hand, scale with both the width and height of the array.
\section{Experiments}
We include three experiments that evaluate the accuracy and area overhead for OverQ.
The first explores the core mechanisms for OverQ to understand how it affects the quantization error and how the quantization error leads to improved accuracy.
The second is an evaluation of OverQ on the accuracy of popular image classification networks evaluated on ImageNet~\cite{deng2009imagenet}. These experiments validate the generality of OverQ across model sizes and quantization techniques.
The improved accuracy has less significance unless the area overhead is sufficiently small over alternative approaches. Therefore, the third experiment quantifies the hardware overhead on an ASIC prototype, showing that the PE additions are insignificant.

 \begin{table*}[tbhp]
  \small
  \centering
  \caption[OverQ ImageNet Evaluation]{\BF{OverQ ImageNet Evaluation --}
  This table evaluates the addition of OverQ on top of popular post-training quantization methods. All results use 8-bit weights and 4 or 5-bit activations, and the OverQ includes a combination of range and precision overwrite with cascading. OCS and ZeroQ are combined with MMSE clipping.
  }
    \vspace{0.1in}
    \begin{tabular}{l|rr|rr|rr|rr}
    \toprule
                & \multicolumn{2}{c|}{\BF{ResNet-18}} & \multicolumn{2}{c|}{\BF{ResNet-50}} & \multicolumn{2}{c|}{\BF{DenseNet-121}} & \multicolumn{2}{c}{\BF{VGG-19}} \\
    \cmidrule{2-9} \BF{Clipping} & \BF{A4   } & \BF{A5    } & \BF{A4   }  & \BF{A5   }  & \BF{A4   } & \BF{A5    } & \BF{A4   }  & \BF{A5   }\\
                   \BF{Method}   &           &            &            &            &           &            &            &          \\
    \midrule
    MMSE    & 67.07  & 68.60 & 46.80 & 57.24 & 72.98  & 74.06  & 71.01 & 73.09 \\
    + OverQ & 68.35  & 68.82 & 51.04 & 58.30 & 73.81  & 74.30  & 71.22 & 73.21 \\ 
    \midrule
    ZeroQ   & 65.40  & 68.50 & 61.49 & 69.41 & 70.36  & 73.30  & 68.01 & 72.51  \\ 
    + OverQ & 67.47  & 68.99 & 65.22 & 70.43 & 73.49  & 73.58  & 71.36 & 73.46 \\
    \midrule
    OCS     & 66.36  & 69.56 & 50.39 & 64.58 & 73.03  & 75.07  & 70.30 & 73.58 \\
    + OverQ & 68.23  & 69.92 & 54.83 & 65.69 & 74.40  & 75.04  & 71.40 & 73.56 \\
    \midrule
    STD     & 64.21  & 69.48 & 68.82 & 76.03 & 70.43  & 74.76  & 71.07 & 73.30 \\
    + OverQ & 69.47  & 71.06 & 74.67 & 76.88 & 75.06  & 76.21  & 71.51 & 73.64 \\
    \bottomrule
    Float   &        & 71.47 &       & 77.72 &        & 76.52  &       & 74.28\\
    \end{tabular}%
  \label{tab:overq-imagenet}%
\end{table*}%

\subsection{OverQ Study}
We first evaluate OverQ across multiple models using post-training quantization techniques. Following the common convention, we do not quantize the first and last layers. Our results do not attempt to reach the state-of-the-art quantization accuracies, which use techniques like dynamic quantization and trained quantization to reach lower bitwidths. These techniques are not easily supported in hardware. Instead, we use only techniques that are widely supported, so that OverQ could be adopted in many hardware designs.

Figure~\ref{fig:overq-std} breaks down how each OverQ component operates and highlights a key tradeoff with uniform quantization. We first profile the activations on a small dataset of 1000 training images to gather the max, min, and std per-channel. We then use the std to sweep the clipping threshold. Std gives an intuitive unit that remains largely unaffected by outliers in the activations. We quantize the weights and activations to 4 bits, using our implementation of asymmetric quantization with additional per-channel weight quantization. Since the systolic array accumulates only within each output channel, our hardware prototype supports per-channel weight quantization.

Figure~\ref{fig:overq-std}(a) exposes the core tradeoff for clipping thresholds. Each method reaches a local maximum accuracy at some clipping threshold, as measured by number of stds from the mean. OverQ with cascading reaches its maximum at the smallest threshold around 3.5 stds, and the baseline reaches it later at 7 stds. The addition of precision overwrite does not affect the position of the maximum since it operates only on non-outliers. Below the maximum too many outliers are clipped, and above the maximum too much quantization error is accumulated on smaller values. 

We analyze these quantization errors in Figure~\ref{fig:overq-std}(b). Here, the values are split between large and small magnitudes, and the error is measured as the total absolute error in a ResNet-18 layer. As the clipping threshold increases, the error increases on the majority of values, which have small magnitudes. Simultaneously, the error on the large outliers decreases. These have opposing behavior because the error on the large values is due primarily to clipping, while the error on the small values is due to the lack of precision. However, with OverQ, this tradeoff is removed for the majority of outliers since they can now overwrite nearby zeros. In Figure~\ref{fig:overq-std}(b), this can be seen with the inclusion of range overwrite, which significantly decreases the error on the large values. Cascading further increases the number of outliers handled and significantly decreases the error to nearly zero. This low error leads to earlier and higher local maximum accuracies for OverQ. In addition, precision overwrite offers additional modest decreases in error for non-outliers by limiting the quantization error due to the lost precision.

\begin{table*}[tbhp]
  \small
    \caption[OverQ Hardware Overhead]{\BF{OverQ Hardware Overhead} -- Area breakdown for an individual PE when modeled in Verilog for range overwrite and full OverQ. The other datapath column includes the additional muxing logic.}
  \centering
    \vspace{0.1in}
    \begin{tabular}{l|ccc}
    \toprule
    \cmidrule{1-4} \BF{Area ($um^2$)} & \BF{Multiply} & \BF{Add} & \BF{Other Datapath} \\
    \midrule
    Baseline & 128.74 & 135.13 & 41.23 \\
    \midrule
    OverQ RO & 128.74 & 141.51 & 80.07 \\
    Overhead  & 0.00\% & 1.36\% & 8.30\% \\
    Overhead +1b & -7.17\% & 0.83\% & 7.37\% \\
    \midrule
    OverQ Full & 128.74 & 141.51 & 88.31 \\
    Overhead & 0.00\% & 1.36\% & 10.06\% \\
    Overhead +1b & -7.17\% & 0.83\% & 8.98\% \\
    Overhead +2b & -13.16\% & 0.57\% & 7.63\% \\
    \midrule
    \end{tabular}%
  \label{tab:overq-area}%
\end{table*}%

\subsection{ImageNet Evaluation}

Table~\ref{tab:overq-imagenet} shows the effects of OverQ across a variety of popular post-training quantization methods and models. For these results, OverQ includes precision overwrite and cascading. We chose a cascade factor of 4 for our experiments, since as shown in Table~\ref{tab:overq-coverage} this value covers the majority of the outliers. Each model is taken from the pytorchcv~\footnote{https://pypi.org/project/pytorchcv/} model zoo, which offers a large variety of cross-framework, pre-trained models for computer vision. We quantize these models with our own framework, which has been internally tested across many projects. When available, we used the open-source implementations of the baseline methods, e.g. ZeroQ and OCS. In this evaluation, we combined ZeroQ and OCS with MMSE clipping. For the STD method, we sweep the clipping threshhold, evaluate on the profiling dataset, and choose the highest accuracy.

Overall, OverQ adds additional accuracy to the baseline results across methods, models, and bitwidths. For example, DenseNet-121 and ResNet-18 add +5\% to their STD baselines at 4 bits, approaching within a few percent of their floating point accuracies. Since STD sweeps the clipping thresholds, it allows OverQ to push to lower thresholds where more outliers are generated. Other methods evaluate at a single analytically-derived clipping threshold that could limit the effects of OverQ. DenseNet-121, for instance, reaches its local maximum accuracy at 9 stds without OverQ and 4 stds with OverQ. Table~\ref{tab:overq-imagenet} shows that STD with OverQ often performs the best across other methods, even achieving floating point accuracies on VGG-19.

The baseline methods are chosen among related works and encompass a variety of post-training and data-free approaches. Accuracy for some methods, especially on ResNet-50, is low due to the low activation bitwidth and the wide activation distributions. With low bitwidths, the tradeoff between outliers and centralized, small values is even more pronounced. MMSE often does not clip aggressively enough at these bitwidths and the accuracy suffers as a result. The baseline results for OCS are comparable to MMSE in most models, likely due to the relatively high weight bitwidth where weight splitting is less effective. Including OverQ on these methods consistently improves their results by ameliorating both the clipping and quantization errors. For instance, at 4 bits OverQ combined with OCS improves ResNet-50 by 4\% and DenseNet-121 by 1\%.

Table~\ref{tab:overq-imagenet} further compares the accuracy of a 4-bit activation to a 5-bit activation for each clipping method to demonstrate the general trend that OverQ offers more benefits at lower bitwidths with more clipping induced outliers. At 5 bits, however, there are fewer outliers and the tradeoff described in Figure~\ref{fig:overq-std} is less pronounced. Therefore, OverQ is most effective when the activation bitwidth is low, suggesting that as 4-bit inference continues to become more commonplace, OverQ becomes even more relevant.

\subsection{Hardware Resources}

The accelerator prototype was written in Verilog and synthesized via the Synopsys Design Compiler.
Table~\ref{tab:overq-area} shows resource usage of the baseline and OverQ designs. This represents one part of the systolic array of the matrix-vector multiply accelerator. The experiments show the main overhead of OverQ is the multiplexer logic at 8.30\% and 10.06\% for OverQ RO and OverQ Full. The second largest overhead is the adder logic at 1.36\% for both OverQ RO and OverQ Full. OverQ does not affect the multiplier datapath. This suggests that a small shift to include SIMD instructions will mitigate the multiplexer overhead, since the larger MAC units will account for a larger portion of the PE. 

Another observation is that providing support for precision overwrite in addition to range overwrite helps to increase accuracy while not affecting the multiplier and add datapaths. Adding cascading on top of that further increases the accuracy while not introducing additional area cost within the PE. These results reflect the core design principle of OverQ to avoid MAC overhead, which is the major area bottleneck of previous hardware solutions for handling outliers such as OLAccel that dedicates a separate PE for sparse outlier computation~\cite{park2018outlier}. OverQ instead leverages the systolic array structure via area efficient storage and muxing logic. This contributes to its scalability for larger designs, as in commercial ML accelerators that increase the systolic array size to up to $256 \times 256$ cells as in Google's TPU~\cite{google2017tpu}.

\section{Conclusions and Future Work}

We introduce and evaluate OverQ as a lightweight hardware-focused technique for improving activation quantization. OverQ opportunistically overwrites nearby values to give extra range or precision during DNN inference. Our experiments demonstrate that OverQ complements existing post-training and data-free quantization techniques. Range overwrite increases the bitwidth for outliers, and when combined with cascading, pushes the outlier coverage to above 90\% on most layers. Precision overwrite adds additional precision to non-outliers. Combined these methods are sufficient to handle the majority of outliers during computation and consistently increase the accuracy on the models tested. For example, ResNet-18 improves by 5\% when quantized with STD. 

These techniques incur small overhead on top of the baseline weight-stationary array. Since the number of PEs scale quadratically with the size of the array, at scale we expect the PE overhead to dominate. Other components of the systems, e.g. the rescaling unit, scale only linearly with the size of the array. Our experiments show that the extra PE area on our ASIC prototype is approximately .50\%, which suggests the total overhead will be limited.

In the future, we hope to expand our hardware prototype to include a full systolic array with quantization, rescaling, routing, and control logic. Using this more advanced prototype, we can explore optimizing the control and routing logic in the rescaling unit to enable a full DNN accelerator. This will create a more versatile accelerator targeted at full model life-cycle.
\subsection*{Acknowledgments}
This work was supported in part by the Semiconductor Research Corporation (SRC) and DARPA. One of the Titan Xp GPUs used for this research was donated by the NVIDIA Corporation. We would also like to acknowledge Shiqi Wang for her contributions and discussions of the project.

\newpage
\bibliography{ms}
\bibliographystyle{icml2021}

\end{document}